\documentclass[runningheads]{llncs}
\usepackage{marvosym}
\usepackage{graphicx}
\usepackage[colorlinks,citecolor=blue,urlcolor=blue,bookmarks=false,hypertexnames=true]{hyperref}
\usepackage{microtype}
\usepackage{graphicx}
\usepackage{subfigure}
\usepackage{booktabs}
\usepackage{longtable}
\usepackage{multirow}
\usepackage{amsmath}
\usepackage{float}
\usepackage[table]{xcolor}
\usepackage{lscape}
\usepackage{booktabs}
\usepackage{color,soul}
\newcommand\blfootnote[1]{%
  \begingroup
  \renewcommand\thefootnote{}\footnote{#1}%
  \addtocounter{footnote}{-1}%
  \endgroup
}

\begin{document}
\title{Recent Trends in Artificial Intelligence-inspired Electronic Thermal Management}
\titlerunning{Artificial Intelligence-inspired Electronic Thermal Management}
\author{Aviral Chharia\inst{1}\textsuperscript{({\large \Letter})}*, Nishi Mehta\inst{2}*, Shivam Gupta\inst{3}*, Shivam Prajapati\inst{4}*}
\authorrunning{Chharia et al.}
\institute{Mechanical Engineering Department, Thapar Institute \\ of Engineering and Technology, Patiala, PB 147004, India \and Mechanical Engineering Department, Sardar Vallabhbhai \\ National Institute of Technology, Surat, India \and Department of Mechanical Engineering, Indian \\ Institute of Technology (ISM), Dhanbad, India \and Department of Mechanical Engineering, National \\ Institute of Technology, Agartala, India \\ \email{\{achharia\_be18\}@thapar.edu}
\\ *Authors claim equal contribution}

\maketitle

\begin{abstract}
The rise of computation-based methods in thermal management has gained immense attention in recent years due to the ability of deep learning to solve complex `physics' problems, which are otherwise difficult to be approached using conventional techniques. Thermal management is required in electronic systems to keep them from overheating and burning, enhancing their efficiency and lifespan. For a long time, numerical techniques have been employed to aid in the thermal management of electronics. However, they come with some limitations. To increase the effectiveness of traditional numerical approaches and address the drawbacks faced in conventional approaches, researchers have looked at using artificial intelligence at various stages of the thermal management process. The present study discusses in detail, the current uses of deep learning in the domain of `electronic' thermal management.


\end{abstract}

\blfootnote{\small{Accepted at the Intl. Conference of Fluid Flow \& Thermal Sciences, Surat, India}}

\section{Introduction}

Thermal management is the technique of controlling temperatures for the enhancement of the performance of the system. One of the recent works in thermal management was by (Zhao et al., 2021). They developed a heat pump system combined with an innovative vehicle thermal management system (VTMS) to compensate for the performance loss caused due to the electric heater in the cabin of Fuel Cell Vehicles. (Moore et al., 2014) investigated the capability of various materials and cubic crystals for the thermal management of electronic devices. (Ghosh et al., 2008) carried out experiments to determine the thermal conductivity of graphene and concluded it to be helpful as a material for thermal management in various nano-electronic circuits. Another work by (Pesaran et al., 2001) carried comparative studies on types of cooling for thermal management of electric and hybrid electric vehicles. (Beircuk et al., 2002) combined Single Wall Nano-Tubes with industrial epoxy to enhance thermo-mechanical properties of the epoxy. They found the material to be valuable for thermal management. (Zweben et al., 1998) explored various composite materials with higher conductivity than aluminum alloys and examined their properties and applications in different areas of thermal management. (Kercher et al., 2003) carried out experimental investigations on microjet technology for cooling electronics and found it more effective than the unforced cooling technology. They further compared it with the standard cooling-fan technology. (Mudawar 2001) explored various advanced cooling methods like jet impingement cooling, detachable heat sinks, etc. (Gallego et al., 2003), developed and experimented on carbon foam for thermal management, and found it more effective than conventional heat sinks. (Kandasamy et al., 2007) experimentally investigated the effectiveness of a Phase Change Material (PCM) package for electronics thermal management. Further, his work numerically confirmed PCMs to be highly suitable for application in transient electronic cooling. (Mallik et al., 2011) worked on finding materials that were more thermally conductive than aluminum, mainly for automotive electronic cooling. (Xin et al., 2014) fabricated large-area freestanding graphene papers with high thermal conductivity of about 1238.3-1434 W m$^{-1}$ K$^{-1}$ with electrospray deposition method combined with continuous roll to roll process.

\begin{figure}[h]
\begin{center}
\centerline{\includegraphics[width=0.7\textwidth]{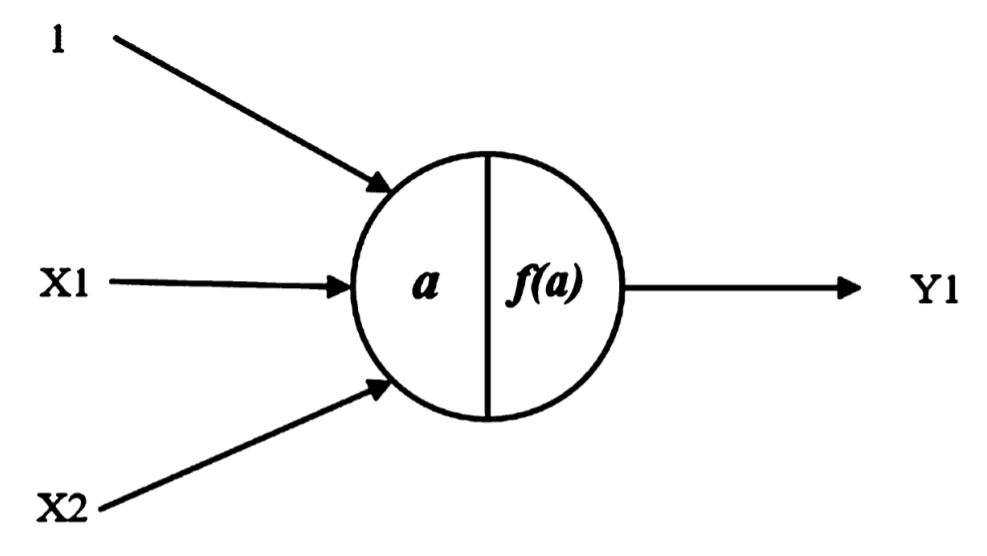}}
\caption{Architecture of the simplest possible Neuron}
\end{center}
\end{figure}

Artificial intelligence (AI) has gotten much attention in recent years because of its vast possibilities. Using AI-based optimization, prediction, etc. Researchers have proposed various new methods that have surpassed the conventional techniques in both accuracy, robustness, and time to be implemented. However, the present literature lacks an in-depth study of the various artificial intelligence-based techniques that have been used in thermal management. In this work, we aim to provide an in-depth review of the recent development of AI-inspired Thermal Management techniques used in various applications like in the area of Single-Core and Multi-Core Processors, high accuracy determination of thermo-physical properties impacting thermal management, predicting the thermal conductivity of composites, system performance prediction accuracy, connection establishing, decision making for the best configuration/method, parameter optimization, etc. We also give an in-depth look at the most current advancements in artificial intelligence approaches used in the thermal management literature.

\section{Recent Advances in Artificial Intelligence In Thermal Management Literature} 

Machine Learning is one of the most rapidly evolving technical disciplines today (Jordan, M. I. and Mitchell, T. M., 2015), and it has been used in various domains, including thermal management. (Mitchell, T. M., 1997) defined Machine Learning as the study of a new class of algorithms that automatically improve their performance over a task `$T$' as measured by the performance metrics `$P$', through experience `$E$'. The last decade has witnessed a rapid rise in Artificial Intelligence algorithms mainly due to the increased computational power of processors.

\subsection{Artificial Neural Networks}

Artificial neural networks, often known as neural networks (NNs), are computing systems inspired by the biological neural networks that make up brains. During training Neural Networks, the main objective is to find a suitable `loss' function and thereafter minimize it through computational techniques. One of the widely used optimization techniques includes stochastic gradient descent (SGD). SGD entails displaying the input vector for several examples, computing the outputs and errors, calculating the average gradient for those samples, and changing the weights as required (LeCun, Y., Bengio, Y. and Hinton, G., 2015). The method is repeated for a large number of small sets of samples from the training set until the loss function's average no longer decreases. Stacking up multiple neural networks forms a deep learning model. The most commonly used loss functions include Binary Cross-Entropy Loss, Mean Square Error, etc.

\subsection{Solving Partial Differential Equations with Physics-Informed Neural Networks}

Recently introduced, Physics Informed Neural Networks (Raissi, M., Perdikaris, P. and Karniadakis, G. E., 2019) are particularly trained to solve supervised learning problems. These form a new class of data-efficient universal function approximators that encode physical laws of nature, naturally as initial input. These are currently employed to obtain data-driven solutions and discover partial differential equations.

\begin{figure}[h]
\begin{center}
\centerline{\includegraphics[width=0.8\textwidth]{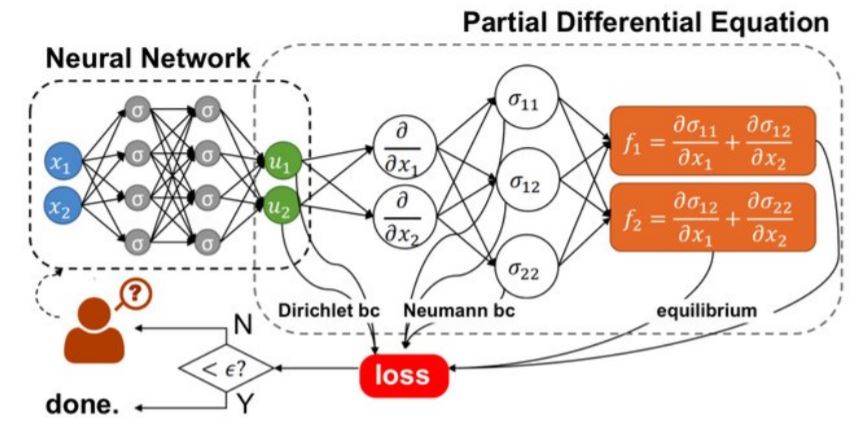}}
\caption{Physics Informed Neural Network (Raissi, M., Perdikaris, P. and Karniadakis, G. E., 2019)}
\end{center}
\end{figure}

\subsection{Convolutional Neural Networks}

Convolutional Neural Networks (CNNs) take high dimensional data as input in the form of matrices, for example, a colored image made of three 2D arrays having pixel intensities in the RGB color channel (Chharia, A. and Upadhyay, R., 2020). The architecture of a CNN is structured one consisting of multiple layers, which includes- the convolutional layers and the pooling layers. A convolutional layer's units are arranged in feature maps, with each unit linked to local regions in the preceding layer's feature maps. The outcome of this locally weighted sum is then processed through a non-linearity (most commonly used is ReLU non-linearity). There have been numerous applications of Convolutional Neural Networks in many different areas of thermal management, as discussed in the subsequent sections. Most authors have used Transfer Learning (Krizhevsky, A., Sutskever, I. and Hinton, G. E., 2017) of pre-trained datasets like ImageNet (Deng, J. et al., 2009). The 3D equivalent of CNNs is 3D CNNs used when the dataset comprises 3D structures/ matrices.

\subsection{Other Deep Learning Architectures and Domains}

Other Deep Learning architectures (Ian Goodfellow and Yoshua Bengio and Aaron Courville, 2016) widely used in thermal management include Long Short Term Memory (LSTM) networks, Recurrent Neural Networks (RNNs), Reinforcement Learning, etc. Table II discusses the in-depth collection of the various recent works in Thermal Management involving Deep Learning and Machine Learning Techniques.

\begin{figure}[h]
\begin{center}
\centerline{\includegraphics[width=\textwidth]{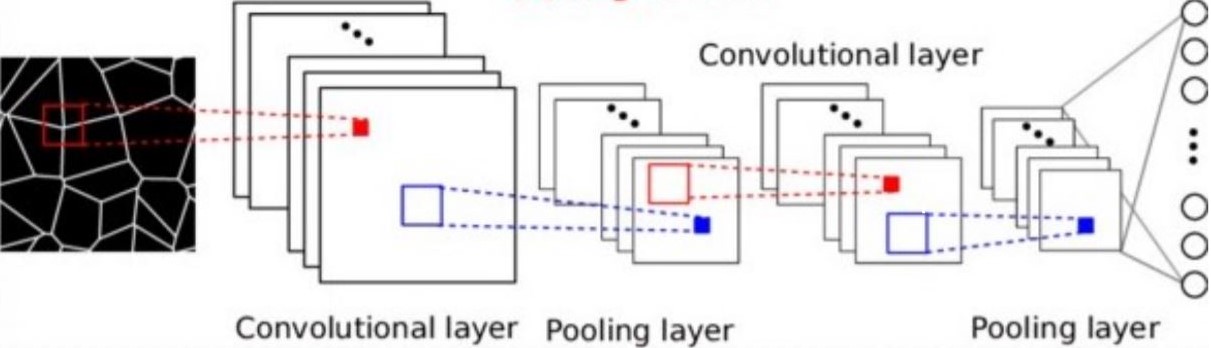}}
\caption{A Convolutional Neural Network with Convolutional and Pooling layers}
\end{center}
\end{figure}

\section{Thermal Management assisted by Machine Learning, Deep Learning and other AI domains}

\begin{center}
    \textbf{Table I.} The percentage increase in various parameters by employed machine learning methods over conventional approaches.
\end{center} 
\begingroup
\begin{longtable}{p{4cm} p{5.5cm}}\toprule
\textbf{Name of Parameter} & \textbf{Percentage increase compared to
Conventional methods}\\
\midrule
\endhead
Video Quality & 17\% increase \\
\midrule
Video performance & 11\% increase \\
\midrule
Average temperature & 12\% decrease \\
\hline
\end{longtable}
\endgroup

\subsection{ML- based optimization of Thermal Design Parameters of Multi-processor systems-on-chip (MPSoCs)}

The enhanced performance of High-Efficiency Video Coding (HEVC) increases the complexity of computational methods. This leads to an increase in the temperature of Multi-Processor Systems-on-Chip (MPSoCs) due to increased power consumption. (Iranfar et al., 2017) proposed an approach for thermal and power management using artificial intelligence. It learned the best encoder configuration and core frequency for every individually running video stream running in the MPSoC. The parameters used for determining them were obtained from frame compression, performance, temperature, and the total power. This approach was further experimented with by implementing it on an enterprise multicore server compared with conventional methods. As shown in Table I, the approach was advantageous as these parameters were improved compared to the conventional techniques. Furthermore, this improvement was achieved with no compromise on power or decrements of compression.

\subsection{Nearest Neighbor Search and ANNs for Improving Transient Performance of Thermal Energy Storage Units}

(Shettigar et al., 2020) improved the transient performance of thermal energy storage units using PCM for latent heat storage by thermal management by applying machine learning. In a phenomenon known as sub-cooling, the solidification of Phase Change Materials (PCM) is initialized by significantly reducing the temperature below the melting point of the PCM. The cold finger technique is used for initiating the solidification of PCM. In this technique, a little unmelted part is kept in the PCM to initialize the remaining PCM's nucleation and solidify. The intended melt fraction of PCM in this technique can be predicted accurately by using the Machine Learning technique, unlike conventional methods, in which the accuracy is increased at the cost of storage capacity. Conventional techniques consisting of physics-based prediction methods had some limitations like-

\begin{enumerate}
    \item High uncertainties in measurement
    
    \item Limited accuracy in real time predictions.
    
    \item Limited operating conditions in which they can be applied
\end{enumerate}

To overcome these limitations, Nearest Neighbour Search (NNS) and Artificial Neural Networks (ANNs) were combined, and the melt fraction of PCM was determined with significantly high accuracy. Therefore, this technique can be considered beneficial during the ending of the process (the stage in which PCM has a higher melt fraction). However, it cannot be considered beneficial for the stage having a lower melt fraction of PCM, based on its accuracy compared to physics-based solvers.

\subsection{3D CNNs for Thermal Management in Composites Through Structure Property Linkage}

Thermal conductivity is an essential parameter in thermal management by composites. Solving partial differential equations (PDEs) and effective medium theory are the conventional methods to determine effective thermal conductivity. The linkage of structure and property could be established through various machine learning methods. Convolutional Neural Networks (CNN) uses deep learning to do generative and descriptive tasks. It uses machine vision having image/video recognition, recommender systems, and natural language processing in order to do so. The thermal conductivity of composites using various CNNs was predicted by (Rong et al., 2019). The three-dimensional convolutional neural networks can more accurately establish the linkage of structure and property. The accuracy of prediction through Machine Learning Methods depends on the microstructure-representing features. The two-dimensional cross-sectional images and two-dimensional convolutional neural networks can very well predict the effective thermal conductivity of three-dimensional composites. Also, they are easier to obtain than three-dimensional microstructure images. Two-dimensional CNNs can predict both isotropic particle-filled and an-isotropic stochastic complex composites by taking 2D images in a proper direction with respect to fillers. Cross-sectional images along the direction of heat flow were found to be more beneficial than cross-sectional images perpendicular to the direction of heat flow. Accuracy was further increased by averaging a large no. of images and using large-size images.

\subsection{Integrating CNNs with Finite Element Method for Thermal Management in Composite Phase Change Materials}

(Kolodziejczyk et al., 2021) determined the thermal properties of composite phase change materials (CPCM) like Paraffin and Copper foam used for thermal management of a Li-ion battery pack by combining CNN with the Finite Element Method (FEM). First, the FEM was used for modeling the microstructure of CPCM. Then, thermal properties were used to create an image dataset, which was further used for training and testing the performance of CNN. Image classification was done by comparison of the dataset using a network architecture. This procedure predicted the thermal characteristics of the CPCMs. They were used in Newman's battery model to simulate the Li-ion battery's cell's heat generation and electrochemical response. Paraffin had the main role in thermal management, in which the copper foam acted as a thermal conductivity enhancer. The thermal management effectiveness of the battery pack was determined by the multi-scale model developed by the use of FEM. Here also, the thermal characteristics which were determined by the combination of CNN and FEM were used. Finally, it was concluded that the combination of CNN with FEM was more accurate for determining the battery thermal management system's effectiveness compared to only FEM.

\subsection{Reinforcement Learning based Thermal Management of High Performing Computing Systems}

With the increase of power demands for high-performance computing systems, this is high time that researchers should focus more on the thermal management of such systems as the CPU. They are designed for very low maximum power rating having minimum performance (D. Brooks and M. Martonosi, 2001) for specific high graphic applications keeping in mind the dynamic management of the control unit. Specific machine learning techniques can be implemented to control the thermal management of multimedia applications' computational intensity. Dynamic Thermal Management (DTM) has also been used in balancing the temperature of different computer systems like servers (Pakbaznia et al., 2010), embedded systems (Zhang, S. and Chatha, K.S., 2010), and general-purpose computers (Cochran, R. and Reda, S., 2010). Work done by (Yang et al.,2011) presented a study on dynamic thermal management using reinforcement learning algorithms and validated their results with a DELL personal computer with Intel Core 2. It works on changing temperature and switching patterns workload by looking at the control unit's temperature sensor and event counters. This learning consists of (R. Sutton and A. Barto, 1998)

\begin{enumerate}
    \item An agent, with a finite action set A.
    
    \item Environment space is represented by S.
    
    \item Policy of $\pi$ representing behaviour of learning agent at any specified time. It can also be defined as the mapping from the set of environment states to the set of actions, i.e. $\pi$: S $\rightarrow$ A
\end{enumerate}

They used dynamic voltage and frequency scaling (DVFS) for controlling the operating temperature. Machine learning finds the easiest way for thermal management, which ultimately helps decrease the high-temperature hotspot, unlike standard DTM methods (Shafik et al., 2016, Rong Ye and Qiang Xu, 2012).

\subsection{Bayesian Learning, Reinforcement Learning and Regression Analysis based prediction for Dynamic Voltage and Frequency Scaling}

Modern multi-core processors suffer higher power densities than previous technology scaling nodes due to the high integration density and various obstructions of voltage scaling. Sometimes these issues might lead to high-temperature spots, which will decrease the overall efficiency and may cause inconsistent aging, increase the chances of chip failures, degrade reliability, and ultimately reduce the system's performance (P.D. et al., 2017). Thus, effective thermal management practices are becoming more relevant than ever. However, the main goal remains to prevent chips from possible overheating while not sustaining unbearable cooling costs, thus guaranteeing to continue technology scaling trends feasible (Pagani et al., 2020). Moreover, minimizing overall energy consumption or energy management while not suppressing the battery lifetime and performance of the system is another significant problem. Overall system performance maximization is typically the most chased optimization goal. However, thermal constraints are the major limiting factors for maximizing performance, especially in modern chips with very high-power densities due to the dark silicon problem (Esmaeilzadeh et al., 2011; Shafique et al., 2014).

Conventional thermal and power management methods largely depend on a certain a-priori understanding of the chip's thermal model and information of the applications to be executed like average and transient power consumption. Nevertheless, this information is hardly available, and if it is, it lacks to reflect the temporal and spatial variations and uncertainties that come from the environment, the hardware, or the workloads. Therefore, the system's thermal manager should consider such uncertainties while power management techniques that are statically adjusted are not likely to achieve the best performance when these characteristics are changing (Wei Liu, Ying Tan, and Qinru Qiu, 2010; Shafik et al., 2016). In contrast, machine learning-based techniques can observe, learn, and adapt to different working environments, making them a potential choice to be employed in varying conditions and workloads. Machine learning (ML) based predictors such as neural networks (Rong Ye and Qiang Xu, 2012), Bayesian learning (Jung and Pedram, 2010; Yanzhi Wang et al., 2013), reinforcement learning (Hantao et al., 2014; Xu et al., 2014, 2018; Lu, Tessier and Burleson, 2015; D. et al., 2016), and regression analysis (Manoj P. D., Yu, and Wang, 2015; Sheng Yang et al., 2015; Sayadi et al., 2017) are also widely utilized for prediction and to perform Dynamic voltage and frequency scaling (DVFS). In addition, ML-based thermal/performance management has the extra benefit of learning the characteristics or trends of workloads to take control of decisions.

\subsection{Q-Learning Algorithm based Linear Adaptation of Lagrangian Multiplier for Minimizing Power in Single-Core Systems}

In (Eui-Young Chung, Benini and De Micheli, 1999), authors propose a DPM technique for a random number of sleep states that shuts down the idle components based on adaptive learning trees and idle period clustering. The proposed methodology correlates with advanced branch prediction schemes of computer architectures to lessen the penalty of mispredicted branches. Also, based on the transition time between sleep states and the expected duration of the next idle period, authors derive a function that selects the optimal sleep state in which to set the core to reduce the energy consumption. In (Tan, Liu and Qiu, 2009; Wei Liu, Ying Tan and Qinru Qiu, 2010), authors propose a DPM technique that minimizes the average power consumption for an arbitrary number of sleep states under a given performance constraint which can alter during runtime. The methodology is based on enhancing the traditional Q-learning algorithm that proposes a linear adaption of the Lagrangian multiplier to hunt for the strategy that minimizes the power consumption while delivering the exact needed performance. Thus, the authors present an iterative algorithm that adjusts the Lagrangian multiplier to the most optimal value for a provided performance constraint.

\subsection{Supervised Learning based Power Management model for energy minimization in Multi-Core Systems}

The work in (Jung and Pedram, 2010) portrays a supervised learning-based power management model for energy minimization on a multi-core chip with per-core DVFS. A probability-based learner is used as the power manager to predict the processor's performance state for every incoming task by examining some commonly available input features and then uses this predicted state to suggest the optimal power management action from a pre-computed policy table. The chief objective of the power manager is to derive a DVFS policy for picking the frequency/voltage levels of the cores that reduces the overall energy consumption based on the load conditions and workload characteristics.

\subsection{Reinforcement Learning based Dynamic Power Management in Multi-Core Systems} 
The work in (Lin, Wang and Pedram, 2016) proposes a reinforcement learning-based DPM structure for data centres that tries to reduce the overall energy utilization of a server pool while sustaining a reasonable average job response time. In (Iranfar et al., 2015), authors developed a machine learning-based thermal management model that employs a heuristic to bound the learning space by assigning a particular set of available actions to all existing states. The purpose of this work is to accelerate the performance while minimizing the thermal stresses under thermal and power constraints by transferring tasks between cores and by selecting the frequency/voltage levels of every core.

\subsection{Discussion and Conclusions}

Artificial Intelligence has been immensely used in Electronic Thermal Management. We provide an in-depth review of the recent development of AI-inspired Thermal Management techniques used in various applications. The present study mainly focuses on the following main aspects of Electronic Thermal Management-

\begin{enumerate}
    \item Various Machine learning algorithms like reinforcement learning algorithms could be used in high graph applications for Dynamic Thermal Management (DTM).

    \item Artificial intelligence techniques can be used for thermal management without compromising the quality of output. Hence they are more effective compared to conventional methods.
    
    \item Next, we have portrayed an overview of various research developments that aim to use Machine Learning and Deep learning to efficiently manage the power and thermal supply of single-core and multi-core processors. 
\end{enumerate}

The chief advantage of using ML techniques is the ability to instantaneously adapt to different system conditions and workloads, learning from past events to improve themselves simultaneously as the environment changes, resulting in highly improved management decisions. By going through these recent applications of AI in thermal management, one can look into the current developments in thermal management techniques and make it more accurate, efficient and time-saving by deploying appropriate AI models.

\pagebreak

\begin{center}
    \textbf{Table II.} Recent works in Thermal Management involving Machine Learning and Deep Learning techniques.
\end{center} 

\begingroup
\begin{longtable}{p{2cm} p{4cm} p{6.5cm}}\toprule

\textbf{Reference} & \textbf{Thermal management Advances} & \textbf{Role of Machine Learning and Deep Learning}\\
\midrule
\endhead

Rong et al., 2018 & Predicting Thermal Conductivity of Composites & Used Convolutional Neural Networks (CNNs) architectures to establish structure and property linkage to determine composite’s thermal conductivity, which controls the performance of thermal management systems.\\
\midrule

Kolodziejczy et al., 2021 & Determined the thermal properties of composite phase change materials (CPCM) like Paraffin and Copper foam used for thermal management of a Li-ion battery pack. & Combined CNN with FEM (which was used for modelling the microstructure of CPCM). Using Thermal properties, an image dataset was created and used for training of CNN. The study concluded that combination of CNN with FEM was more accurate in battery thermal management system’s effectiveness determination compared to FEM alone.\\
\midrule

Yang et al., 2011 & Dynamic Thermal Management of Multimedia application using reinforcement learning & The main aim of this algorithm is to maximize the average long-term applications by trial-and-error interaction with a dynamic environment. It is developed by $\pi$ learning policy which is a mapping between states and actions. This technique is used to control the optimum temperature of any multimedia application using machine learning.\\
\midrule

Iranfar et al., 2017 & Thermal management of Multi-Processor Systems on Chips & Approach for selection of best encoder configuration and core frequency was developed using artificial intelligence for temperature maintenance of MPSoCs.\\
\midrule

Zhang et al., 2017 & Machine learning is used to predict temperature for runtime thermal control across system components. & 1-Feature selection techniques were utilised to increase the performance of machine learning methods that were previously created. 2-A thorough comparison evaluation of prediction approaches using network and linear regression-based methods. 3- It is simpler to estimate the temperature of a component with a minimum error of 2.9°C and 3.8°C using machine learning approaches such as Lasso linear regression.\\
\midrule

Shettigar et al., 2020 & Improvement in transient thermal performance of thermal energy storage units by application of machine learning & Nearest Neural Search and Artificial Neural Network were used to determine PCM melt fraction very much accurately, which was an important parameter controlling the thermal performance.\\
\midrule

Tang et al., 2021 & Performance of battery thermal management system in electric vehicles using machine learning. & Particle swarm optimization (PSO) algorithm was used to to optimise the thermal management of the battery.\\
\midrule

Rong et al., 2019 & Deep learning was used to predict the composite's thermal conductivity from a section picture. & Convolutional neural networks (CNNs) are used to extract geometric features of the components which helps in establishing structure property linkages.\\
\midrule

Chung et. al, 1999 & Proposal of new DPM scheme based on adaptive learning trees and idle period clustering & A comprehensive algorithm based on an Adaptive Learning Tree is applied to binary and multi-valued sequences to predict the future idle periods with high accuracy.\\
\midrule

Liu et al., 2010 & Enhanced Q-learning algorithms-based novel power management techniques & Model free online algorithms are presented for dynamic power management with performance constraints. Using Machine Learning algorithms, it can adapt itself to the changing performance constraint during runtime.\\
\midrule

Jung et. al, 2010 & Energy minimization on multi-core chips using supervised learning & A probability-based learner is built using supervised learning to estimate the processor's performance state for every incoming task and thus using the former knowledge to minimize energy requirement.\\
\midrule

Lin et. al, 2016 & Efficiently managing energy utilised by data centres through reinforcement learning & A DPM-based on reinforcement learning. \\
\hline

\end{longtable}
\endgroup

\end{document}